\documentclass{article}
\usepackage[left=3.5cm,right=2.5cm,top=2cm,bottom=2cm]{geometry}
\usepackage[T1,T2A]{fontenc}
\usepackage[utf8]{inputenc}
\usepackage[english]{babel}
\usepackage[dvipsnames]{xcolor}

\usepackage{longtable}
\usepackage{setspace}
\usepackage{hyperref}
\usepackage{todonotes}
\usepackage{cite}
\usepackage{paratype}
\usepackage{paralist}
\usepackage{microtype}
\usepackage{booktabs}
\usepackage{lscape}
\usepackage{graphicx} 

\usepackage{authblk}

\usepackage{paralist}
\usepackage{url}
\usepackage{setspace}
\usepackage{longtable}
\usepackage{caption}
\usepackage{cite}
\usepackage{multirow}
\usepackage{amsmath,amssymb}

\title{HJ-Ky-0.1: an Evaluation Dataset \\ for Kyrgyz Word Embeddings}
\date{}
\begin{document}

\author[1,2,3,4]{Anton M. Alekseev}
\author[4]{Gulnara Kabaeva}

\affil[1]{Steklov Institute of Mathematics at St. Petersburg, \newline St. Petersburg, Russia}
\affil[2]{St. Petersburg State University, St. Petersburg, Russia}
\affil[3]{Kazan (Volga Region) Federal University, Kazan, Russia}
\affil[4]{KSTU named~after I.~Razzakov, Bishkek, Kyrgyzstan}

\maketitle

\begin{abstract}
    One of the key tasks in modern applied computational linguistics is constructing word vector representations (word embeddings), which are widely used to address natural language processing tasks such as sentiment analysis, information extraction, and more. To choose an appropriate method for generating these word embeddings, quality assessment techniques are often necessary. A standard approach involves calculating distances between vectors for words with expert-assessed ``similarity''. This work introduces the first ``silver standard'' dataset for such tasks in the Kyrgyz language, alongside training corresponding models and validating the dataset’s suitability through quality evaluation metrics\footnote{This is a translation of the original paper published in 2023, in Russian. In case the reported results or any outcomes of the study find some use in your work, we suggest that you cite it as follows: \texttt{Alekseev, A., Kabaeva, G.: HJ-Ky-0.1: an Evaluation Dataset for Kyrgyz Word Embeddings. Herald of KSTU 68(4) (2023)}, \\ or, in bibtex: \texttt{@article\{alekseev2023embeddings, author=\{Alekseev, Anton and Kabaeva, Gulnara\}, title=\{\{HJ-Ky-0.1: an Evaluation Dataset for Kyrgyz Word Embeddings\}\}, journal=\{Herald of KSTU\}, volume=\{68\}, number=\{4\}, year=\{2023\}, url=\{https://www.elibrary.ru/item.asp?id=58483453\}\}}.}.

\end{abstract}

\section{Introduction}\label{introduction}

Natural language processing (NLP) is a multidisciplinary field at the intersection of applied linguistics, artificial intelligence, machine learning, and statistics. Automatic text processing methods play a crucial role in modern data science since valuable information is often presented as unstructured, free-form text rather than in machine-readable formats. Despite the wealth of research in computational linguistics, particularly in morphology (e.g.,~\cite{bakasova2016}), there is currently a lack of open datasets and language resources for processing the Kyrgyz language~\cite{mirzakhalov2021turkic}. It is difficult to envision further progress in this area without addressing this gap.\footnote{An incomplete list of~open tools and datasets can be found, e.g.,~at~the~link \url{https://github.com/alexeyev/awesome-kyrgyz-nlp}.} Specifically, no datasets are available for evaluating semantic methods, including those designed to assess the quality of word vector representations (word embeddings).

The goal of this work is to begin filling this gap by introducing the first dataset tailored for evaluating word vector quality in Kyrgyz. This dataset was created through manual translation from Russian. We describe the dataset (pairs of words and their estimated ``similarity'' scores), the translation methodology, and computation of quality metrics for several word embedding models. Furthermore, we validate the dataset's relevance to the task through these quality evaluation measures. The metrics used include Spearman’s rank correlation and Pearson’s correlation coefficient, comparing the original similarity scores with the similarity scores of trainable vector representations (measured by cosine similarity).
This paper provides a brief overview of non-contextual word vector representation methods, evaluation approaches, and details of both the original and translated datasets, including translation strategies. We report the experimental setups, including hyperparameters and descriptive statistics for the datasets. Finally, we summarize the results and discuss potential shortcomings of the dataset, along with steps for improving it and advancing vector semantic methods for the Kyrgyz language.

\section{Building vector representations}\label{sec:embeddings_intro}

Word, subword, and text embeddings have become essential components of many modern natural language processing methods. A key property of these embeddings is that the distance (Euclidean, cosine, etc.) between vectors should be smaller for words with closer ``senses''. Most of these approaches rely on the principle of distributional semantics: ``You shall know a word by the company it keeps!''~\cite{firth1957synopsis}. In other words, words that appear in similar contexts often share certain properties, not limited to syntax. Zellig Harris~\cite{harris1954distributional} illustrated this with the words ``oculist'' and ``eye-doctor'', which, being synonyms, frequently co-occur with terms such as ``eye'' or ``examined''.

Early methods for constructing word vectors were based on term-document matrices and their factorized modifications, such as Latent Semantic Indexing (LSI)~\cite{deerwester1988improving}. Other methods utilized term-term (or term-context) matrices (e.g., a ``vocabulary'' $\times$ ``vocabulary'' matrix based on word co-occurrence frequencies within a window of consecutive words), often weighted using Pointwise Mutual Information (PMI). These approaches essentially involved calculating and re-weighting large sets of statistics.

A significant breakthrough in vector representation quality was achieved in 2013 with predictive methods introduced in \emph{word2vec} models~\cite{mikolov2013efficient,mikolov2013distributed}. These models use shallow neural networks to predict a target word based on its surrounding context (Continuous Bag-of-Words, CBOW) or to predict the context based on a target word (Skip-Gram). The hidden layer values of the trained network are then used as word embeddings. Later, the \emph{fastText} model~\cite{bojanowski2017enriching,grave2018learning} was developed to address vocabulary dependence and preprocessing challenges, particularly for morphologically rich languages like Kyrgyz, which lack high-quality context-aware lemmatizers. In \emph{fastText}, each word is represented as a combination of character n-grams, summing these to compute the final vector representation. This approach is less sensitive to vocabulary and word normalization. The predictive approach of \emph{word2vec} (specifically the Skip-Gram with Negative Sampling variant) was later shown to implicitly factorize a PMI-based term-context matrix~\cite{NIPS2014_feab05aa}. Other notable non-contextual embeddings include GloVe~\cite{pennington2014glove}, which incorporates word frequencies, and \emph{dict2vec}~\cite{tissier2017dict2vec}, which uses dictionary definitions as training data. Additionally, Paragraph2vec (doc2vec)~\cite{le2014distributed}, inspired by \emph{word2vec}, was developed for document-level representations, though newer and more efficient methods now exist for representing short texts.

Since 2013, trainable vector representations have been successfully applied to numerous NLP problems across diverse domains~\cite{tutubalina2017using,filchenkov2017morpheme}. However, since the emergence of contextual embeddings in late 2017, models based on recurrent neural networks~\cite{peters-etal-2018-deep} or the Transformer architecture~\cite{NIPS2017_3f5ee243,devlin-etal-2019-bert} have demonstrated superior performance by incorporating contextual information during both training and inference. Despite this, non-contextual models remain crucial for tasks without explicit context or for languages with limited resources, such as Kyrgyz. These models, along with \emph{fastText} and \emph{word2vec}, are also commonly used as baselines for word and short text representations.

\section{The original dataset}\label{sec:dataset}

The HJ~\cite{panchenko2015russe} dataset is a collection of Russian noun pairs derived from translating the well-known English-language WordSim353~\cite{agirre2009study} dataset, along with the RG~\cite{rubenstein1965contextual} and MC~\cite{miller1991contextual} datasets, into Russian. Word similarity judgments were newly obtained through crowdsourcing, where annotators rated 15 randomly selected word pairs (out of 398 total pairs) on a scale from 0 (not related at all) to 3 (high similarity) using a specialized interface. Details of the annotation process are described in the original work~\cite{panchenko2015russe}.

In total, $4{,}200$ ratings were gathered from $280$ marking sessions. The consistency of the scores was evaluated using Krippendorff’s alpha, which resulted in a value of $0.49$. The published dataset includes each word pair along with the average score from the annotators, normalized to fall within the range $[0, 1]$.

\section{Translation into the Kyrgyz language}\label{sec:translation}

During the process of translating the HJ dataset into Kyrgyz, several challenges arose, necessitating the development of specific rules. These challenges stemmed from the following issues: 
\begin{inparaenum}[(1)]
\item words in any language can have multiple meanings (senses), and in Kyrgyz, each sense may correspond to a distinct word (e.g., the Russian word <<лук>> can mean ``onion'' (<<пияз>>) or ``bow'' (<<жаа>>) depending on the context);
\item even when there is no semantic ambiguity, a word may have more than one translation; and 
\item for semantic similarity evaluation, it is necessary to translate the original Russian word into a single Kyrgyz word, which is not always feasible.
\end{inparaenum}
The rules we followed are detailed below.

By default, we used K. K. Yudakhin's Kyrgyz-Russian dictionary~\cite{yudakhin1957russko} for translation, facilitated by a web interface provided by the creators of \url{el-sozduk.kg}. If the source word was semantically ambiguous (e.g., «лук»), the translation closest in meaning to the second word in the pair was selected, based on the assumption that annotators likely scored the original dataset intuitively in a similar manner. If a word was not found in Yudakhin's dictionary, other resources were consulted, with preference given to terminological and spelling dictionaries~\cite{law2014,termin2007,ortho2009karasaev,juri2005,enky2015}, followed by E. D. Asanov's dictionary~\cite{asanov_dictionary}, and finally others. Comments in the dataset indicate the sources used for each word pair.

If no single-word translation could be found for one or both words in a pair, the pair was excluded from the dataset. When choosing between a direct lexical borrowing from Russian and an alternative translation, preference was given to the alternative. If Yudakhin's dictionary~\cite{yudakhin1957russko} only provided a lexical borrowing, but an alternative consistent with other rules was found in terminological dictionaries or Asanov's dictionary, the latter was used. Proper names not found in dictionaries were translated based on relevant articles from the Kyrgyz Wikipedia (\url{https://ky.wikipedia.org/}).

As a result, 361 word pairs were created, each paired with similarity scores from the HJ dataset. The dataset and accompanying code will be made publicly available after the publication of this work and potentially after a word embeddings competition\footnote{\url{https://github.com/alexeyev/kyrgyz-embedding-evaluation}}. A sample of the data is shown in Table~\ref{tab:data_example}.

\begin{table}[ht]
    \centering
    \begin{tabular}{|c|c|l|}
        \hline
        \textbf{Word $1$} & \textbf{Word $2$} & \textbf{Similarity}\\ 
        \hline
        \multicolumn{3}{|c|}{...} \\
        \hline 
        иерусалим (jerusalem) & ысрайыл (israel) & 0.6222 \\
        шайман (instrument) & курал (weapon) & 0.6222  \\
        планета (planet) & жылдыз (star) & 0.6191 \\
        өлкө (country) & жаран (citizen) & 0.6191 \\
        жолборс (tiger) & фауна (fauna) & 0.6191 \\
        студент (student) & профессор (professor) & 0.6191 \\ 
        \hline
        \multicolumn{3}{|c|}{...} \\
        \hline 
        марс (mars) & суу (water) & 0.0909 \\
        азчылык (minority) & дүйнө (world) & 0.0909 \\
        кылым (century) & улут (nation) & 0.0889 \\ 
        падыша (king) & капуста (cabbage) & 0.0889 \\
        багуу (care) & архитектура (architecture) & 0.0889 \\
        \hline
        \multicolumn{3}{|c|}{...} \\
        \hline 
    \end{tabular}
    \caption{Pair samples from the constructed HJ-Ky-0.1 dataset.}
    \label{tab:data_example}
\end{table}

\section{Word embeddings}

The prepared collection of word pairs is specifically designed to evaluate non-contextual word vector representations. To obtain preliminary results and verify the adequacy of the quality estimates, we utilized pre-trained word embeddings for both Kyrgyz and Russian languages. Additionally, we trained several models using publicly available Kyrgyz text collections, as described below.

\paragraph{Pre-trained \emph{fastText} and Compressed \emph{fastText} embeddings.}

The fastText models for $157$ languages, including Kyrgyz and Russian, were released online alongside the original publication~\cite{grave2018learning}. These models were trained using the Continuous Bag-of-Words (CBOW) scheme (see Section~\ref{sec:embeddings_intro}) with a window size of $5$, $10$ negative samples, and a vector dimension of $300$. The models were trained on language-specific segments of the CommonCrawl dataset.

However, the models obtained in this manner are very ``heavy'' in terms of the number of parameters and overall required memory, and the usage of the software that employs them e.~g. on mobile devices and laptop computers is limited due to the high requirements for the computer’s RAM capacity. Specialized compression methods have been developed to solve this problem by reducing the size of the model without significant losses in quality. For example, matrix decomposition via SVD (singular value decomposition), vector quantization~\cite{jegou2010product}, ``re-design'' of the hashing trick used in fastText, feature selection, etc.

However, these models are resource-intensive in terms of memory and computational requirements, making them less suitable for use on mobile devices or computers with limited RAM. To address this issue, specialized compression techniques, such as singular value decomposition (SVD), vector quantization~\cite{jegou2010product}, and enhanced hashing tricks, have been developed to reduce model size without significant quality loss. These methods, along with options for combining them, are implemented in the {\tt avidale/compress-fasttext} library~\cite{dale_compress}. Compressed versions of the fastText models were published by Liebl Bernhard~\cite{liebl_bernhard_2021_4905385}. For our experiments, we used the relevant compressed models: {\tt fasttext-ky-mini} for Kyrgyz and David Dale's {\tt geowac\_tokens\_sg\_300\_5\_2020-100K-20K-100.bin} model\footnote{\url{https://github.com/avidale/compress-fasttext/releases/download/gensim-4-draft/geowac_tokens_sg_300_5_2020-100K-20K-100.bin}} for Russian, based on the corresponding model~\cite{KutuzovKuzmenko2017}.

\paragraph{Training word embeddings on Leipzig Corpus data.}

We also trained embeddings using data from the Leipzig Corpus Collection\footnote{\url{https://corpora.uni-leipzig.de/}}~\cite{goldhahn2012building}, focusing on the largest available NewsCrawl segment ({\tt kir\_newscrawl\_2016\_1M}, one million sentences) and Wikipedia segment ({\tt kir\_wikipedia\_2021\_300K}\footnote{\url{https://wortschatz.uni-leipzig.de/en/download/Kirghiz}}, 300,000 sentences). Sentences were tokenized using a tokenizer based on Apertium-Kir~\cite{washington2012finite}, which was also used as a stemmer. The stemming process reduced tokens to their first morphological segment as analyzed by Apertium-Kir. While this approach results in some information loss, it was necessary due to the absence of a context-aware lemmatizer for Kyrgyz.

For our first baseline, we used the Skip-Gram Negative Sampling (SGNS) variant of \emph{word2vec}, as implemented in the \emph{gensim}-4.2.0 library~\cite{rehurek_lrec}. We trained vectors of dimensions 100 and 300, with a window size of 5, 5 negative samples per positive example, and 10 training epochs over the corpus.

Next, we trained \emph{fastText} embeddings on the Leipzig Corpus data. The training scheme was also Skip-Gram Negative Sampling, with 10 epochs, vector dimensions of 100 and 300, a window size of 5, and 10 negative samples. Character n-grams of 3 to 6 characters and 2,000,000 hashing buckets were used for the hashing trick.

It is worth noting that we chose SGNS for training because it is considered better suited for relatively small datasets with a large number of rare words compared to CBOW, as suggested by one of the \emph{word2vec} authors\footnote{\url{https://groups.google.com/ g/word2vec-toolkit/c/NLvYXU99cAM/m/E5ld8LcDxlAJ}}. The hyperparameters were chosen based on commonly used values in research and practice to ensure that the results would reflect the HJ-Ky-0.1 dataset’s overall suitability rather than being influenced by specific parameter tuning.

\section{Results}\label{sec:results}

The results of the data preparation and model training are summarized in Table~\ref{tab:scores_results}, which presents the quality scores achieved by the models.

\begin{table}[ht]
    \centering
    \begin{tabular}{|c|c|c|c|c|c|c|} \hline
    Model family & Training scheme & Data & Preprocessing & dim & $r_s$ & $\rho$ \\ \hline
    word2vec & SGNS & Leipzig & Apertium tokenization &  100 & 0.513 & 0.458 \\    
    word2vec & SGNS & Leipzig & Apertium tokenization &  300 & 0.524 & 0.472 \\
    word2vec & SGNS & Leipzig & Apertium stemming &  100 & 0.440 & 0.383 \\
    word2vec & SGNS & Leipzig & Apertium stemming &  300 & 0.448 & 0.403 \\ \hline    
    fastText & SGNS & Leipzig & Apertium tokenization & 100 & \textbf{0.605} & \textbf{0.572} \\
    fastText & SGNS & Leipzig & Apertium tokenization & 300 & 0.601 & 0.560 \\ 
    fastText & SGNS & Leipzig & Apertium stemming & 100 & 0.583 & 0.571 \\ 
    fastText & SG & Leipzig & Apertium stemming & 300 & 0.588 &  0.564 \\ 
    \hline
    fastText & CBOW & CC-Ky~\cite{grave2018learning} & Europarl tokenization & 300 & 0.557 & 0.515 \\ 
    C-fastText & CBOW & CC-Ky~\cite{grave2018learning} & Europarl tokenization & 100 & 0.480 & 0.457 \\ \hline
    fastText & CBOW & CC-Ru~\cite{grave2018learning} & Europarl tokenization & 300 &0.245 & 0.302 \\ 
    C-fastText & CBOW & CC-Ru~\cite{grave2018learning} & Europarl tokenization & 100 & 0.243 & 0.283 \\ 
    \hline
    \end{tabular}
    \caption{Quality estimates achiveved with pre-trained models and models trained on Leipzig corpus. Here \textbf{dim} is the vector size, $r_s$ and $\rho$ are the Spearman's and Pearson's correlation coefficients, respectively.. C-fastText are Compressed fastText representations. ``Europarl tokenization'': \url{https://www.statmt.org/europarl/}.}   
    \label{tab:scores_results}
\end{table}

The purpose of this work is not to compare models but to present a new dataset and indirectly validate its suitability for processing Kyrgyz text, as it was constructed through manual translation. Several observations support this validation.

1. \emph{Performance differences by language}. Despite the presence of lexical borrowings from Russian in the Kyrgyz dataset, Russian word embeddings trained on much larger text corpora performed significantly worse, indicating that the dataset is well-aligned with the specific characteristics of the Kyrgyz language.

2. \emph{Impact of stemming}. The ``rough'' stemming approach resulted in reduced quality scores, which is expected since this method causes significant information loss. For \emph{word2vec}, stemming even led to the absence of representations for some words.

3. \emph{Model comparison}. The fastText models consistently outperformed word2vec, reflecting expectations for morphologically rich languages like Kyrgyz. The performance drop from tokenization to stemming is particularly notable, as certain words are excluded from the dataset after stemming.

4. \emph{Corpus quality}. The fastText model trained on the Leipzig Corpus (fastText, Skip-Gram, Leipzig, 100) outperformed the CommonCrawl-Ky model, even without hyperparameter tuning. This suggests that the Leipzig Corpus is cleaner and more suitable than the automatically collected CommonCrawl dataset, which may include mislabeled or noisy text.

These results indicate that the HJ-Ky-0.1 dataset provides reliable quality metrics for word embeddings. However, further validation on a manually re-annotated dataset is necessary to confirm these findings. We also note t hat it might be preferable to train models on text data in the Kyrgyz language, since the model published by \emph{Facebook} (CBOW, CC-Ky, 300, $r_s = 0.557$, $\rho = 0.515$) might not be the best available (which, again, should be confirmed by testing on a manually annotated set of pairs).

\section{Conclusion}\label{sec:conclusion}

In this work, we introduce the first word vector quality evaluation dataset for the Kyrgyz language, HJ-Ky-0.1, created by translating the HJ dataset from Russian. We provided indirect evidence of the dataset’s adequacy for the task, including:
\begin{itemize}
    \item superior performance of Kyrgyz word embeddings compared to Russian embeddings on the Kyrgyz dataset;
    \item a decline in quality when using stemmed text, demonstrating the limitations of rough stemming for morphologically rich languages;
    \item better results from \emph{fastText} compared to \emph{word2vec}, which aligns with expectations for Kyrgyz text processing.
\end{itemize}

Our experiments also showed that \emph{fastText} embeddings trained on the Leipzig Corpus outperform those trained on CommonCrawl, highlighting the importance of clean, high-quality text data for training. These findings suggest that the HJ-Ky-0.1 dataset is a valuable resource for selecting word representation methods and fine-tuning model hyperparameters.

Looking ahead, we plan to improve the dataset by manually re-annotating the word pairs with expert ratings provided by native speakers, as translation may have altered the perceived similarity of some pairs. We also aim to expand the dataset with additional pairs, potentially derived from synonym dictionaries. These enhancements will facilitate more reliable evaluations and support the development of high-quality vector semantic methods for the Kyrgyz language.

\section*{Acknowledgments}
The authors express their gratitude to S. I. Nikolenko for insightful comments and for proofreading both versions of this paper.

\bibliographystyle{plain}
\bibliography{98_references}

\end{document}